\documentclass[12pt]{article}

\usepackage{geometry}
\usepackage{authblk} 
\usepackage[hidelinks]{hyperref}
\usepackage{graphicx} 
\usepackage{amsmath, amssymb} 
\usepackage{booktabs} 
\usepackage{enumitem} 
\usepackage{natbib} 
\usepackage{abstract} 
\usepackage[ruled,vlined]{algorithm2e}
\usepackage{caption}
\usepackage{subcaption}
\usepackage{float} 
\usepackage{newtxtext,newtxmath} 
\usepackage{microtype} 

\title{
    \hrule height 3pt \vspace{10pt}
    \textbf{\Large Exploring the Landscape for Generative Sequence Models for Specialized Data Synthesis}
    \vspace{10pt} \hrule height 1pt
}

\author[1]{\textbf{Mohammad Zbeeb}}
\author[2]{\textbf{Mohammad Ghorayeb}}
\author[3]{\textbf{Mariam Salman}}
\affil[1,2,3]{\small Department of Electrical and Computer Engineering, American University of Beirut}
\affil[1,2,3]{\texttt{\{mbz02, mrg11, mcs12\}@mail.aub.edu}}

\date{}

\begin{document}

\maketitle

\renewcommand{\abstractnamefont}{\normalfont\bfseries}
\renewcommand{\abstracttextfont}{\normalfont\small}

\begin{abstract}
\noindent \normalsize Artificial Intelligence (AI) research often aims to develop models that can generalize reliably across complex datasets, yet this remains challenging in fields where data is scarce, intricate, or inaccessible. This paper introduces a novel approach that leverages three generative models of varying complexity to synthesize one of the most demanding structured datasets: Malicious Network Traffic. Our approach uniquely transforms numerical data into text, re-framing data generation as a language modeling task, which not only enhances data regularization but also significantly improves generalization and the quality of the synthetic data. Extensive statistical analyses demonstrate that our method surpasses state-of-the-art generative models in producing high-fidelity synthetic data. Additionally, we conduct a comprehensive study on synthetic data applications, effectiveness, and evaluation strategies, offering valuable insights into its role across various domains. Our code and pre-trained models are openly accessible at \href{https://github.com/Moe-Zbeeb/Exploring-the-landscape-for-generative-models-for-specialized-data-generation.git}{Github}, enabling further exploration and application of our methodology.  \\  
\textit{Index Terms}: Data synthesis, machine learning, traffic generation, privacy preserving data, generative models.
\end{abstract}

\section{Introduction}
Machine learning algorithms depend heavily on the availability and quality of training data. However, acquiring real-world data poses challenges due to privacy concerns, limited accessibility, and potential biases \citep{lu2024machinelearningsyntheticdata}. Consequently, synthetic data generation has attracted increasing interest, aiming to create diverse and representative datasets that mitigate issues of data scarcity, bias, and privacy \citep{Abadi_2016}.

In recent years, Generative Adversarial Networks (GANs) have emerged as a powerful technique for producing realistic synthetic data \citep{goodfellow2014generative}. GANs are widely applied in fields such as image generation, network traffic modeling, and healthcare data synthesis \citep{antoniou2018dataaugmentationgenerativeadversarial}. These models replicate the statistical properties of real-world data, providing a valuable tool for augmenting datasets in cases where data is limited or sensitive \citep{flowGan}.

Despite their impressive results, GANs face challenges. Computational complexity and training instability have been widely documented, complicating replication across domains \citep{borji2019pros}. Moreover, GANs’ primary focus on unstructured data raises questions about their suitability for structured numerical data, which is often critical in fields such as cybersecurity, finance, and healthcare \citep{flowGan}. This has fueled demand for alternative generative models capable of efficiently handling structured data while preserving the original data’s key statistical properties.

Beyond GANs, Variational Autoencoders (VAEs) and other generative models have shown promise for synthetic data generation. VAEs effectively capture complex data distributions in recommendation systems and collaborative filtering \citep{liang2018variationalautoencoderscollaborativefiltering}. However, they may lack representational power compared to GANs, especially with complex datasets \citep{antoniou2018dataaugmentationgenerativeadversarial}.

Alongside methodological advancements, several studies have integrated privacy-preserving mechanisms into generative models. Differentially private GANs, for instance, generate synthetic data that maintains privacy and minimizes the risk of sensitive information leakage \citep{Abadi_2016}. Such approaches are essential in sensitive domains like healthcare, where data privacy is paramount, requiring a careful balance between data quality and ethical considerations.

Existing synthetic data generation methods often focus on unstructured data or encounter challenges in specialized fields like cybersecurity and financial risk modeling. Furthermore, adversarial training and continuous distribution modeling can complicate the generation process, particularly for structured numerical data with irregularities or outliers \citep{flowGan}.

This paper builds on existing research by exploring the potential of sequence models for synthetic data generation. Sequence models, widely used in natural language processing, present a novel approach to generating structured data by framing it as a language modeling problem \citep{bengio}. By leveraging sequence models’ strengths in handling both discrete and continuous data, we aim to address limitations posed by traditional generative models, particularly in domains requiring structured, high-dimensional data.

Informed by key findings from the literature, we aim to contribute to the discourse on synthetic data generation by investigating how sequence models can provide a computationally efficient alternative to established techniques. While GANs and VAEs have dominated the field, we propose that sequence models offer a flexible and scalable approach for generating high-quality synthetic data, particularly in scenarios with structured data and categorical variables \citep{vaswani2023attentionneed}.

\section{Techniques}
Here we describe our techniques for data generation and training the generators as language-based classifiers. First, we provide background on structured datasets and the data used in our experiments.

\subsection{Dataset Overview}
An overview of the data used in our experiments is presented in Table~\ref{tab:dataset}.

\begin{table}[H]
    \centering
    \caption{Overview of Typical Attributes in Flow-Based Data \citep{flowGan}}
    \label{tab:dataset}
    \begin{tabular}{|c|l|l|l|}
        \hline
        \textbf{\#} & \textbf{Attribute}               & \textbf{Type}         & \textbf{Example}             \\ \hline
        1  & Date First Seen           & Timestamp        & 2018-03-13 12:32:30.383 \\ \hline
        2  & Duration                  & Continuous       & 0.212                   \\ \hline
        3  & Transport Protocol        & Categorical      & TCP                      \\ \hline
        4  & Source IP Address         & Categorical      & 192.168.100.5            \\ \hline
        5  & Source Port               & Categorical      & 52128                    \\ \hline
        6  & Destination IP Address    & Categorical      & 8.8.8.8                  \\ \hline
        7  & Destination Port          & Categorical      & 80                       \\ \hline
        8  & Bytes                     & Numeric          & 2391                     \\ \hline
        9  & Packets                   & Numeric          & 12                       \\ \hline
        10 & TCP Flags                 & Binary/Categorical & .A..S.                  \\ \hline
    \end{tabular}
\end{table}

The dataset used in this project contains attributes typical of unidirectional NetFlow data \citep{flowGan}. NetFlow data is highly heterogeneous, containing continuous, numeric, categorical, and binary attributes. Most attributes, such as IP addresses and ports, are categorical. Additionally, there is a timestamp attribute (Date First Seen), a continuous attribute (Duration), and numeric attributes such as Bytes and Packets.

A key aspect of the dataset is the inclusion of TCP flags, defined here as binary/categorical. These flags can be interpreted either as six binary attributes (e.g., isSYN flag, isACK flag) or as a single categorical value, allowing flexibility in data processing and modeling across generative approaches. This diverse mix of attribute types poses challenges for synthetic data generation, especially when aiming to preserve the statistical properties of the original data while maintaining categorical, continuous, and binary relationships.

\subsection{Data Transformation via CICFlowmeter-V4.0 (ISCXFlowMeter)}
For our experiments, we converted raw network traffic data into CSV format using CICFlowmeter-V4.0, formerly known as ISCXFlowMeter. CICFlowmeter is a bi-directional flow generator and analyzer for Ethernet traffic, specifically designed for anomaly detection in cybersecurity datasets \citep{flowGan}.

CICFlowmeter has been extensively used in well-known cybersecurity datasets, including:

\begin{itemize}
    \item Android Adware-General Malware dataset (CICAAGM2017),
    \item IPS/IDS dataset (CICIDS2017),
    \item Android Malware dataset (CICAndMal2017), and
    \item Distributed Denial of Service (CICDDoS2019).
\end{itemize}

By leveraging CICFlowmeter, we extracted 80 features from each flow, compiling a comprehensive set of flow-based features in CSV format. This structured tabular data includes attributes such as source and destination IP addresses, transport protocols, port numbers, byte and packet counts, TCP flags, and other network metrics.

The conversion of network flows into structured tabular data is crucial for our approach, as it allows for systematic analysis and modeling. The resulting dataset, with its rich set of 80 features, provides the necessary structured format for advanced machine learning techniques, enabling effective synthetic data generation while preserving relationships between features. Structured data in this format facilitates the use of sequence models and other generative techniques that rely on well-organized, tabular data representations \citep{ioffe2015batchnormalizationacceleratingdeep}.

\begin{figure}[H]
    \centering
    \includegraphics[width=0.7\textwidth]{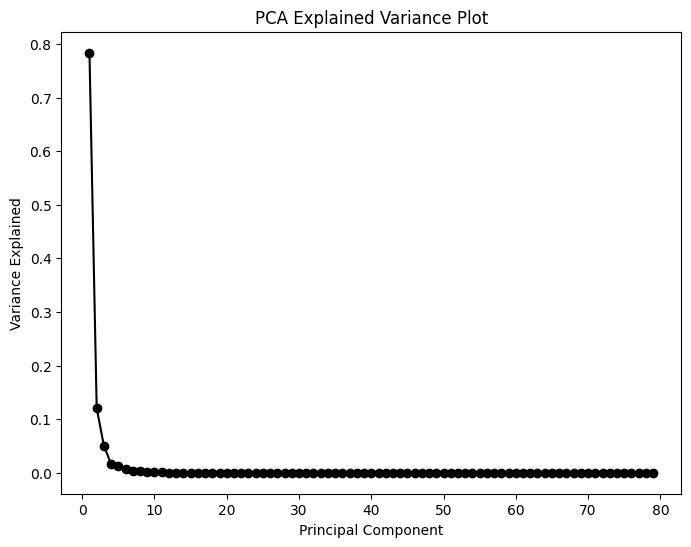}
    \caption{PCA Explained Variance Plot: The majority of the variance is captured by the first few principal components, indicating that much of the data's complexity can be explained by a small subset of components.}
    \label{fig:pca_plot}
\end{figure}

\subsection{Data Transformation to Text Domain - Symbolic Encoding}
Our preliminary exploratory data analysis revealed significant complexity within the dataset, as evidenced by high variance in certain features and the considerable number of unique values across columns. This complexity makes the dataset unsuitable for traditional statistical sampling or simple data synthesis techniques, supporting the exploration of advanced synthetic data generation models.

To enhance the representational quality and address the challenges posed by this complexity, we applied a novel encoding strategy, transforming the dataset from a numeric to a symbolic, textual domain. Specifically, each numeric feature was discretized into intervals, with each interval represented by one of 49 unique symbols. Each symbol corresponds to a 1\% range of the respective feature’s values, resulting in a robust dataset of 30,000 examples. Each example can be considered analogous to a sentence in the symbolic domain \citep{vaswani2023attentionneed}.

This transformation repositions the data generation task as a classification problem rather than continuous regression. By encoding each data point as a sequence of symbols, we frame the task as the prediction of the next symbol in a sequence, given a preceding set of symbols, analogous to language modeling tasks in NLP \citep{bengio}. The dataset, now framed in a discrete symbolic space, facilitates the use of classification algorithms designed for categorical outputs, aligning well with sequence models.

\begin{algorithm}[H]
\caption{Symbolic Encoding Strategy for Dataset Transformation}
\SetAlgoLined
\KwIn{Numerical dataset $\mathcal{D}$ with $N$ elements}
\KwOut{Transformed dataset $\mathcal{D'}$ consisting of 30,000 examples, each represented as a sequence of symbols}

\textbf{Initialize}: Define a set $\mathcal{S}$ of 49 distinct symbols, each representing a 1\% interval of the data range\;
\textbf{Divide}: Partition the range of each numerical feature into 49 equal intervals corresponding to the symbols in $\mathcal{S}$\;

\ForEach{element $e \in \mathcal{D}$}{
    Determine the interval to which $e$ belongs\;
    Map $e$ to the corresponding symbol $s \in \mathcal{S}$\;
}

\textbf{Augment}: For each sequence of data points, prepend a designated start symbol to indicate the beginning of a packet\;

\Return $\mathcal{D'}$: Transformed dataset as sequences of symbols\;
\end{algorithm}

\subsection{Problem Framing}
Our study frames the data generation task as the prediction of the next symbol in a sequence, given the current token. Let $x$ represent the current token, and $y$ the next symbol to be predicted. The probability mass function (PMF) for the random variable $y$, conditioned on $x$, is given by $P(y | x)$, where $P(y | x)$ represents the probability of the next symbol $y$, given the current token $x$. Our goal is to maximize $P(y = y_{\text{true}} | x)$, where $y_{\text{true}}$ is the true label of the next token.

We frame this task as a classification problem, not a regression problem. Although one might bypass text transformation by regressing the output directly, regression introduces challenges, especially when managing high-dimensional, continuous outputs with complex data structures \citep{he2015delvingdeeprectifierssurpassing}.

Classification, by contrast, allows the model to discretize decision-making and capture the data’s discrete nature effectively. In cases where classes occupy distinct manifolds within the data space, classification models can partition the space, yielding probabilistic predictions and clearer boundaries.

\begin{figure}[H]
    \centering
    \includegraphics[width=0.8\textwidth]{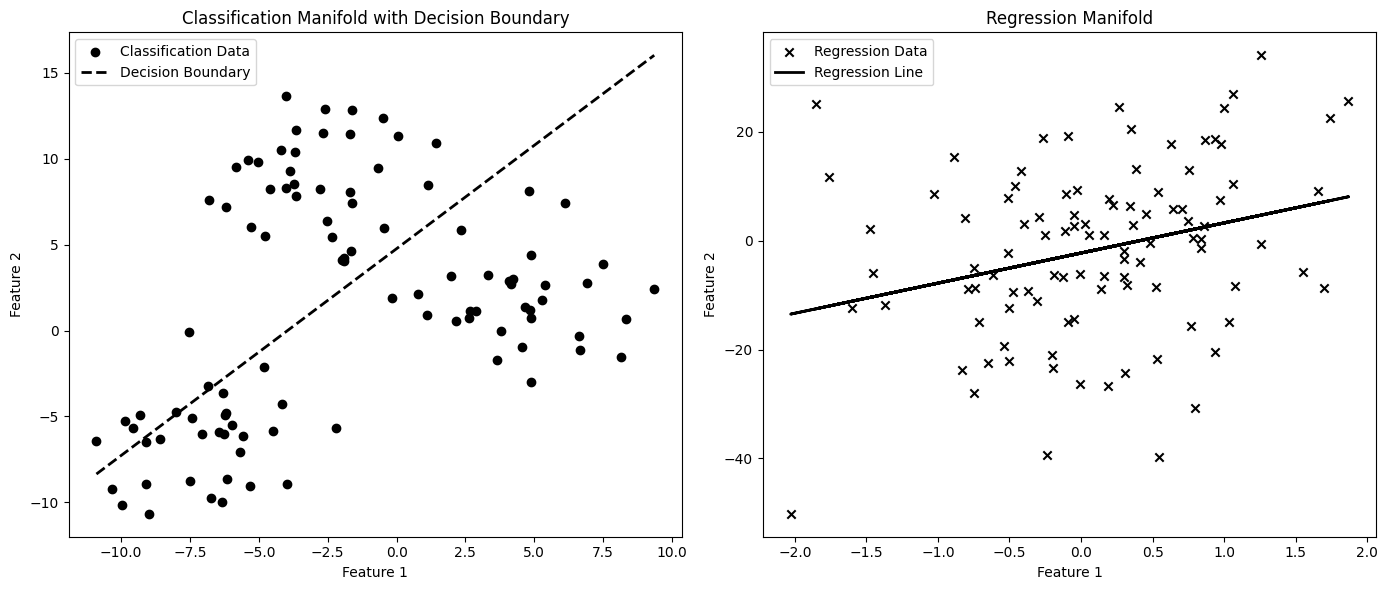}
    \caption{Comparison of Classification and Regression Manifolds. The left plot represents the classification problem with a decision boundary, while the right plot shows the regression problem with a fitted regression line.}
    \label{fig:classification_regression}
\end{figure}

\subsection{Overview of Sequence Models Employed in Our Study}
\subsubsection{WaveNet-Enhanced Neural Probabilistic Language Model}

We employed the WaveNet architecture to enhance a neural probabilistic language model, leveraging its capability to capture intricate sequential dependencies within data. This integration advances language modeling for synthetic data generation. Neural probabilistic language models, initially introduced by Bengio et al. \citep{bengio}, learn distributed token representations and predict sequences based on contextual probabilities. By integrating the WaveNet architecture, developed by Google \citep{DBLP:journals/corr/OordDZSVGKSK16}, we extend this foundational approach.

WaveNet’s use of \textbf{causal convolutions} ensures temporal consistency in predictions—essential for modeling sequential data tasks. The architecture predicts each token based on preceding context, enabling effective capture of linguistic structure and nuances.

\[
y_t = f(x_{t-k}, x_{t-k+1}, \dots, x_t) = \sum_{i=0}^{k} w_i \cdot x_{t-i}, \quad \text{for } t \geq k
\]

\begin{figure}[H]
    \centering
    \begin{subfigure}[t]{0.48\textwidth}
        \centering
        \includegraphics[width=\textwidth]{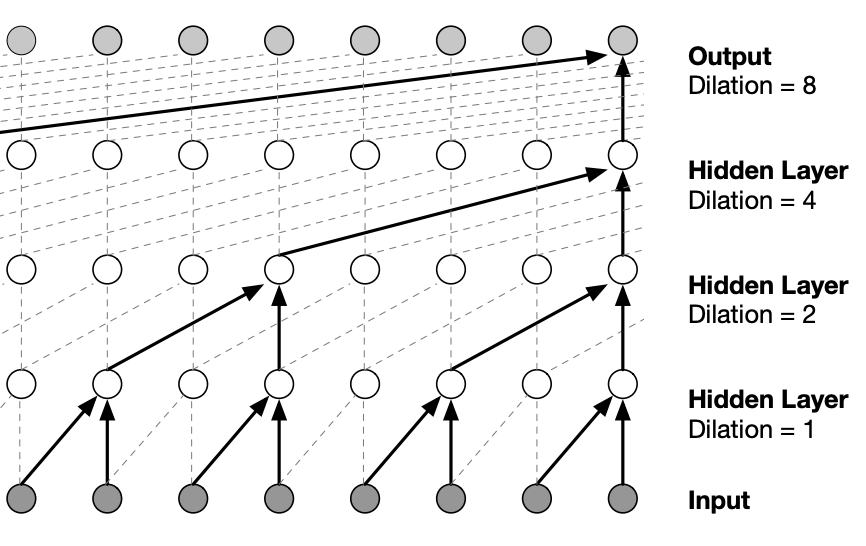}
        \caption{WaveNet Architecture \citep{DBLP:journals/corr/OordDZSVGKSK16}}
        \label{fig:wavenet}
    \end{subfigure}
    \hfill
    \begin{subfigure}[t]{0.48\textwidth}
        \centering
        \includegraphics[width=\textwidth]{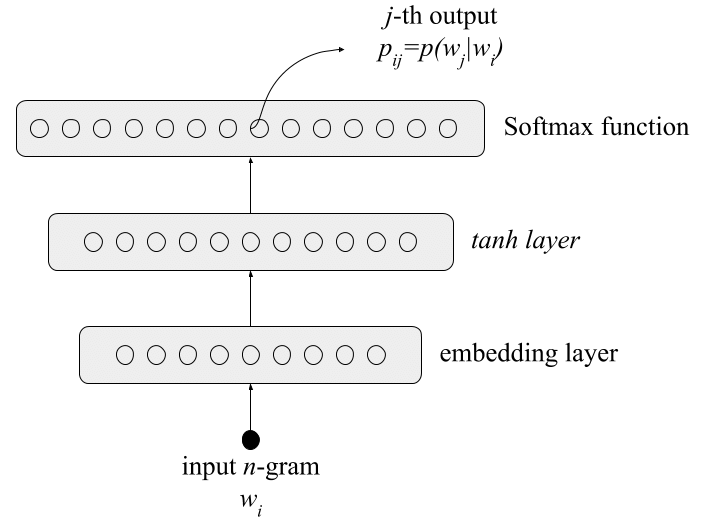}
        \caption{Neural Probabilistic Language Model with Causal Convolutions}
        \label{fig:Neural Probabilistic Language Model with Causal Convolutions}
    \end{subfigure}
    \caption{Architectures Used in Our Study}
    \label{fig:architectures}
\end{figure}

\subsubsection{Recurrent Neural Networks (RNNs)}

Recurrent Neural Networks (RNNs) effectively process sequential data by maintaining a "memory" of previous inputs, achieved through feedback loops in the architecture. This enables RNNs to learn sequence patterns and relationships, producing coherent and contextually relevant text. Our RNN architecture leverages these capabilities, processing encoded data sequences and capturing dependencies within each 10-character segment.

\begin{figure}[H]
    \centering
    \includegraphics[width=0.7\linewidth]{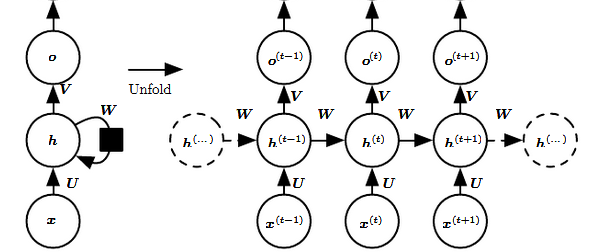}
    \caption{A Recurrent Neural Network}
    \label{fig:recurrent_neural_network}
\end{figure}

\subsubsection{An Attention-Based Decoder - Transformer}

The Transformer \citep{vaswani2023attentionneed} sets itself apart from traditional neural networks by avoiding recurrent mechanisms and instead leveraging self-attention, which weighs the importance of different tokens in an input sequence in parallel. This enables efficient parallel processing and better handling of long-range dependencies.

Our Transformer architecture employs an embedding layer with size 64 per symbol, followed by 4 Transformer blocks, each with 4 attention heads, capturing patterns in sequential data. Each Transformer block includes multi-head attention, feed-forward networks, and layer normalization, supporting robust learning of input sequences.

\begin{figure}[H]
    \centering
    \includegraphics[width=0.7\linewidth]{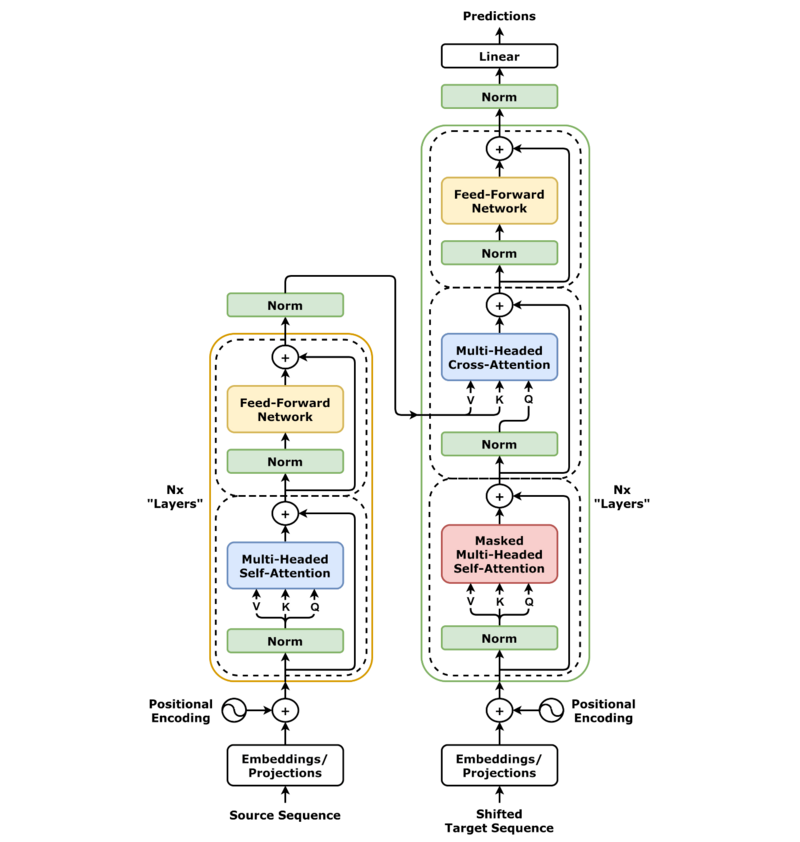}
    \caption{Transformer Architecture}
    \label{fig:transformer_layer}
\end{figure}

\section{Experiment Setup - Framework for Generating Synthetic Data}

This section outlines the models used to create a novel framework for generating synthetic data. We detail the rationale behind selecting these models, discuss the appropriate loss functions, and highlight best practices in training for optimal performance. Additionally, we examine trade-offs involved in generating synthetic data, focusing on aspects of realism, diversity, and privacy preservation.

\subsection{Building Intuition}

The proposed framework is based on the concept of N-gram models \citep{Ngrams}. It involves sampling from a distribution where each character is characterized by a conditional probability over the previous $n-1$ characters.

Mathematically, this is represented as:

\begin{equation}
P(c_i \mid c_{i-(n-1)}, \ldots, c_{i-1})
\end{equation}

This approach has limitations, such as failing to capture long-range dependencies and contextual semantics. Additionally, as $n$ increases, the number of possible N-grams grows exponentially, leading to data sparsity and many zero-count N-grams if the training data is insufficient.

Our approach builds upon Bengio's work on Neural Probabilistic Language Models, where he proposed a neural network architecture to learn the probability distribution of word sequences \citep{bengio}. By integrating these ideas with the WaveNet architecture, known for its strong performance in modeling long-range dependencies in sequential data, we aim to develop a powerful language model capable of generating highly realistic and diversified synthetic text data \citep{DBLP:journals/corr/OordDZSVGKSK16}.

To clarify our methodological choices, we provide intuition behind adopting Bengio's neural network approach and emphasize its advantages in our context.

Bengio's neural network represents each word with a sampled vector and feeds it into a neural network that predicts the next word in the sequence. This prediction is achieved by turning the logits into a distribution via the softmax function, allowing sampling from this distribution. The network learns both the network parameters and the sampled distribution.

Building on this, we introduce WaveNet. Bengio’s approach squashes the input through the network, making it difficult to learn long-term dependencies and positional information. In contrast, WaveNet heavily relies on dilated causal convolutions, specifically designed to capture long-term dependencies by applying multiple large dilated convolutions in parallel.

Mathematically, a dilated convolution operation for a sequence $x$ with filter $f$ is defined as:

\[
y(t) = \sum_{k=0}^{K-1} f(k) \cdot x(t - r \cdot k)
\]

where:

\begin{itemize}
  \item $y(t)$ is the output at time step $t$,
  \item $K$ is the filter size,
  \item $f(k)$ represents the filter weights,
  \item $x(t - r \cdot k)$ are the input values with dilation rate $r$.
\end{itemize}

Using dilated convolutions, WaveNet can efficiently model dependencies over much longer sequences. This is achieved by applying multiple layers of dilated convolutions in parallel, with exponentially increasing dilation rates \citep{DBLP:journals/corr/OordDZSVGKSK16}. This allows the network to capture a broader context at each layer, effectively modeling long-term dependencies.

WaveNet not only models bigrams but also higher-order n-grams (e.g., fourgrams) by progressively squashing the input through these convolutional layers. This gradual reduction in dimensionality captures more semantics, resulting in more realistic and contextually aware synthetic data generation \citep{DBLP:journals/corr/OordDZSVGKSK16}.

Moving to other language models for synthetic data generation, we include the well-known Recurrent Neural Network (RNN). In an RNN, we hold a state $h_t$ and pass it to the forward neuron to maintain contextual information. Mathematically, this is represented as:

\[
h_t = \sigma(W_h h_{t-1} + W_x x_t + b)
\]

where:

\begin{itemize}
  \item $h_t$ is the hidden state at time step $t$,
  \item $W_h$ and $W_x$ are weight matrices,
  \item $x_t$ is the input at time step $t$,
  \item $b$ is the bias,
  \item $\sigma$ is the activation function (e.g., $\tanh$ or $\text{ReLU}$).
\end{itemize}

Next, we examine the Transformer model. In the Transformer, we maintain key, query, and value vectors. The self-attention mechanism in Transformers can be represented as \citep{vaswani2023attentionneed}:

\[
\text{Attention}(Q, K, V) = \text{softmax}\left(\frac{QK^T}{\sqrt{d_k}}\right) V
\]

where:

\begin{itemize}
  \item $Q$ is the query matrix,
  \item $K$ is the key matrix,
  \item $V$ is the value matrix,
  \item $d_k$ is the dimension of the key vectors.
\end{itemize}

To enhance vector representation, we add the original vector to the value vector. Once character representation is well-learned, we can stack a probabilistic model on top of the Transformer. Even simpler models can effectively predict the next character in the sequence. For example, we can use a simple probabilistic model such as a softmax layer:

\[
P(c_i \mid c_{<i}) = \frac{\exp(z_i)}{\sum_{j} \exp(z_j)}
\]

where $P(c_i \mid c_{<i})$ represents the probability of character $c_i$ given the previous characters $c_{<i}$, and $z_i$ is the logit for character $c_i$.

\subsection{Loss}

For generative tasks, where we predict the next character in a sequence from a distribution, cross-entropy loss is commonly used. This loss measures the difference between the true distribution and the predicted distribution for each packet (sequence) in the dataset.

To compute the loss over an entire sequence of packets, we sum the cross-entropy loss over all characters (or time steps) within each sequence, and then average the loss over all sequences in the dataset. The cross-entropy loss for the dataset can be defined as:

\[
\mathcal{L}_{\text{cross-entropy}} = - \frac{1}{M} \sum_{j=1}^{M} \sum_{i=1}^{N} \sum_{c=1}^C y_{i,j} \log(\hat{y}_{i,j})
\]

where:

\begin{itemize}
  \item $M$ is the total number of sequences (or packets) in the dataset,
  \item $C$ is the number of classes (characters or possible outputs for each token),
  \item $N$ is the number of time steps in a sequence,
  \item $y_{i,j}$ is the true probability of class $i$ in sequence $j$ (typically 0 or 1),
  \item $\hat{y}_{i,j}$ is the predicted probability of class $i$ in sequence $j$.
\end{itemize}

This loss formulation ensures that the model sums the loss over all time steps in each sequence, then sums over all sequences in the dataset, and finally averages the loss by the number of sequences $M$.

\subsection{Training Practices}

Generative models require additional care to ensure they produce high-quality and realistic synthetic data. Our framework includes best practices to address these needs effectively.

To address $\tanh$ issues in Bengio's approach, we reference \cite{he2015delvingdeeprectifierssurpassing}. During the forward pass, the activations passing through the $\tanh$ layer tend to be extreme, often lying on the tails at either positive one or negative one.

During the backward pass, when neurons with $\tanh$ activation function update their weights, they often encounter a zero gradient. Consequently, in the update step:

\[
\frac{\partial L}{\partial a} \leftarrow \frac{\partial L}{\partial a} + \frac{\partial L}{\partial y_j} U_j,
\]

the neuron behaves in a shut-off mode due to a zero gradient, resulting in no weight change.

To address this, we manage the standard distribution of activations entering the $\tanh$ activated layer to have a gain of $\frac{5}{3}$ over $\sqrt{\text{fan in}}$, allowing the neurons to learn normally.

For covariant shift resulting from high-dimensional datasets (curse of dimensionality), we apply batch normalization \citep{ioffe2015batchnormalizationacceleratingdeep} as best practice for normalizing the flow (backward and forward) \citep{wu2021rethinkingbatchbatchnorm}. For a layer with $d$-dimensional input $\mathbf{x} = (x^{(1)}, \ldots, x^{(d)})$, we normalize each dimension $x^{(k)}$ as follows:

\[
x^{(k)}_{\text{norm}} = \frac{x^{(k)} - \mathbb{E}[x^{(k)}]}{\sqrt{\text{Var}[x^{(k)}]}}
\]

where:

\begin{itemize}
  \item $x^{(k)}$ is the $k$-th dimension of the input,
  \item $\mathbb{E}[x^{(k)}]$ is the expected value (mean) of $x^{(k)}$,
  \item $\text{Var}[x^{(k)}]$ is the variance of $x^{(k)}$.
\end{itemize}

By normalizing each dimension of the input, batch normalization mitigates covariant shift effects, improving network stability and performance during training.

To address high initial loss in classification generative tasks, we scale output weights during initialization by a small value ($\epsilon$), allowing similar probabilities across alphabets during the first pass.

Let $W_{\text{out}}$ be the weight matrix connecting the hidden layer to the output layer, and let $U_{\text{out}}$ be the bias vector at the output layer, initialized as:

\[
W_{\text{out}} \sim U(-\epsilon, \epsilon),
\]
\[
U_{\text{out}} \sim U(-\epsilon, \epsilon),
\]

where $U(-\epsilon, \epsilon)$ is a uniform distribution. This initialization strategy ensures that weights and biases are scaled by $\epsilon$, yielding a roughly uniform probability distribution for output activations.

\[
W_{\text{out}}(i, j) \sim U(-\epsilon, \epsilon), \forall i, j,
\]
\[
U_{\text{out}}(i) \sim U(-\epsilon, \epsilon), \forall i.
\]

\begin{figure}[H]
    \centering
    \begin{subfigure}[t]{0.45\textwidth}
        \centering
        \includegraphics[width=\linewidth]{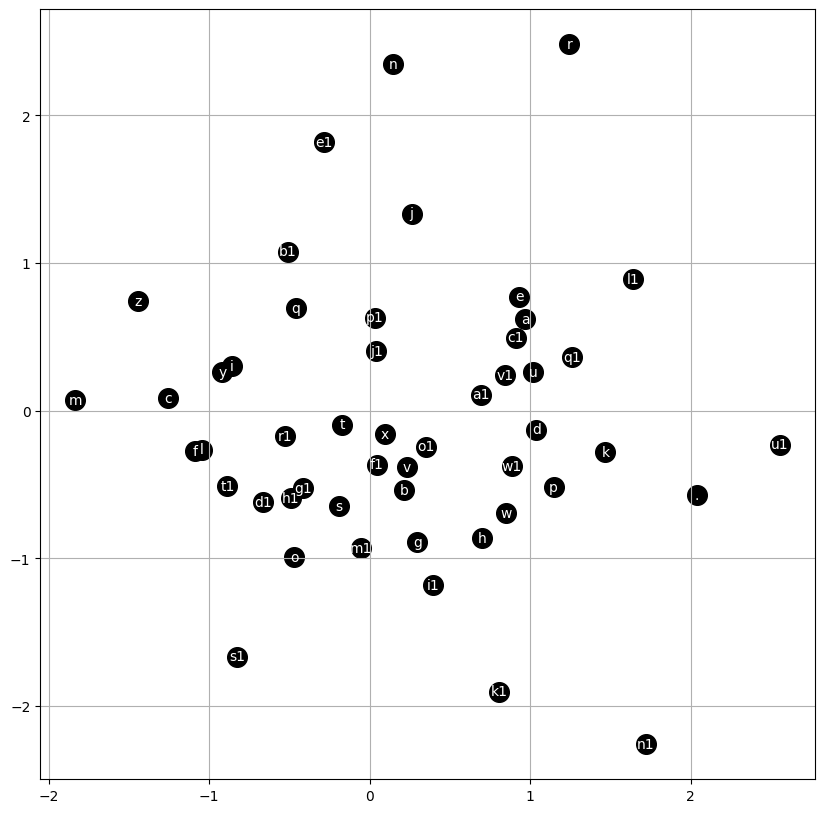}
        \caption{Latent Space}
    \end{subfigure}
    \hfill
    \begin{subfigure}[t]{0.45\textwidth}
        \centering
        \includegraphics[width=\linewidth]{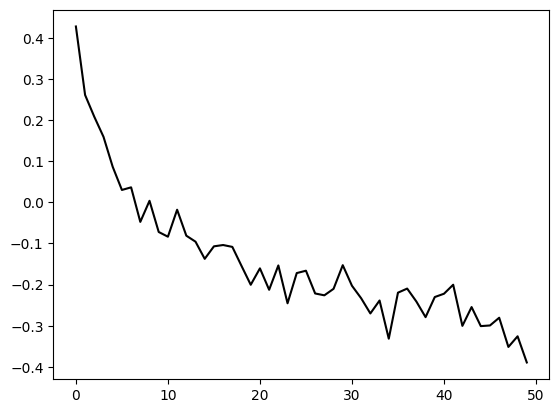}
        \caption{General Decay of the Loss}
    \end{subfigure}
    \caption{Post Learning State - Symbols being learnt and Loss decay}
    \label{fig:post_learning}
\end{figure}

\section{Statistical Framework for Testing Generative Examples}

We posit that if the distribution of generated data closely aligns with the real data distribution, it should effectively train machine learning models. To assess this, we can train a separate classifier on real data to evaluate the statistical validity of the generated data. Logically, training machine learning systems on synthetic data that closely mirrors real data should not harm performance. Overfitting typically arises when a model is overly complex relative to the dataset size, often occurring when there are more parameters than data points.

However, our generated data mitigates this risk by significantly expanding the dataset, providing a more robust foundation for model training. Mathematically, with an original dataset size of $N$ and generated data size $M$, the total dataset size becomes $N + M$. Ensuring that the generated data adheres to the original distribution, $P_{\text{real}}(x) \approx P_{\text{gen}}(x)$, helps prevent model memorization of specific examples, promoting the learning of generalizable patterns \citep{sallab2019unsupervisedneuralsensormodels}. Consequently, incorporating generated data enhances the model’s generalization to unseen examples, rather than leading to overfitting.

Underfitting, which occurs when a classifier fails to capture underlying patterns in the data, resulting in suboptimal performance, is effectively addressed by the generated data. By training on a broader range of examples, the model can better recognize diverse features and gain a comprehensive understanding of the underlying data patterns.

Thus, we employed a one-class Support Vector Machine (SVM) with a linear kernel to determine whether the generated data is statistically similar to real data.

\begin{figure}[H]
    \centering
    \includegraphics[width=0.9\linewidth]{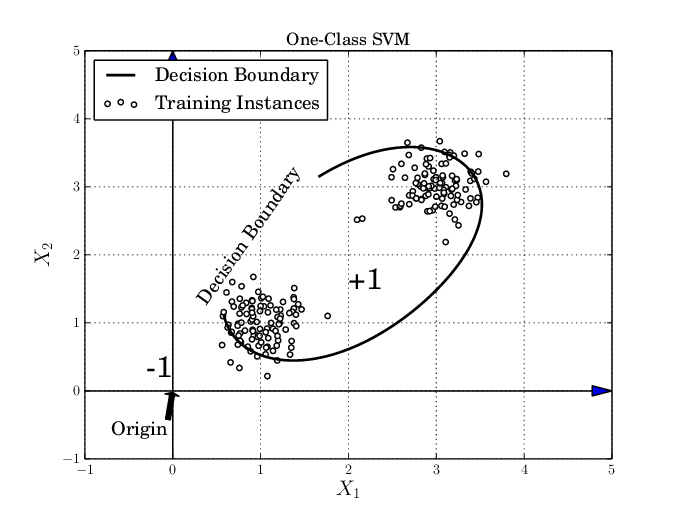}
    \caption{Pseudo Visualization of The Latent Space Post Classifying Inliers}
    \label{fig:svm_visualization}
\end{figure}

\section{Results}
In our experiments, we evaluated each model's ability to generate synthetic data that closely aligns with the original data distribution. The primary evaluation metric was the percentage of inliers, defined as the proportion of generated data points that fall within the distribution of the original data.

The results indicate that while all models performed well, the Recurrent Neural Network (RNN) achieved the highest percentage of inliers at 87.9\%, followed by the Transformer-based Decoder at 84.9\%. WaveNet, although effective in modeling long-range dependencies, had the lowest inlier rate at 69.2\%, likely due to its convolutional structure, which may not capture certain complex dependencies as efficiently as the RNN and Transformer models.

\begin{table}[htbp]
\centering
\caption{Inliers with Respect to Each Model}
\label{tab:results}
\begin{tabular}{|l|c|}
\hline
\textbf{Model}          & \textbf{Inliers (\%)} \\ \hline
WaveNet                 & 69.2\%                \\ \hline
RNN                     & 87.9\%                \\ \hline
Transformer Decoder     & 84.9\%                \\ \hline
\end{tabular}
\end{table}

The RNN outperformed the other models in terms of generating inliers, likely due to its ability to capture sequential dependencies in the data. However, as datasets grow more complex, particularly with higher dimensionality or heterogeneity, the Transformer-based Decoder model is expected to excel. This is due to the Transformer’s self-attention mechanism, which is particularly suited for handling complex dependencies and long-range interactions, which become more significant with increased data complexity.

While WaveNet is designed to model long-range dependencies through dilated convolutions, it may not have been as effective for this dataset due to its convolutional architecture, which can limit its capacity to capture fine-grained patterns in structured data \citep{DBLP:journals/corr/OordDZSVGKSK16}. Nevertheless, its performance might improve with further fine-tuning and optimization.

\section{Synthetic Data Generation: A Survey}

Synthetic data generation has emerged as a vital solution in artificial intelligence (AI) and machine learning, offering unique advantages for both research and practical applications. In response to growing privacy concerns and limited access to real-world data, synthetic data has evolved as a powerful alternative, enabling model training, testing, and deployment without compromising sensitive information. This survey examines the diverse applications of synthetic data generation, from vision and voice technologies to business intelligence, and highlights its potential to transform data-driven fields. By synthesizing insights from recent studies, this survey aims to provide a comprehensive overview of how synthetic data is revolutionizing AI across various domains while addressing privacy and ethical considerations.

\section{Applications}

Synthetic data presents numerous compelling benefits, making it a highly attractive option across a wide range of applications. By streamlining the processes of training, testing, and deploying AI solutions, synthetic data enables more efficient and effective development. Furthermore, this cutting-edge technology mitigates the risk of exposing sensitive information, thereby safeguarding customer security and privacy. As researchers transition synthetic data from the laboratory to practical implementations, its real-world applications continue to expand. This section examines several notable domains where synthetic data generation substantially impacts addressing real-world challenges.

\subsection{Vision}
Generating synthetic data for computer vision tasks has proven highly effective, as it allows for the creation of large, diverse datasets that can be used to train models without the need for costly and time-consuming data collection efforts \citep{marwala2023usesyntheticdatatrain}. These synthetically generated datasets can capture a wide range of scenarios, including complex lighting conditions, occlusions, and diverse object appearances, which are crucial for developing robust vision-based systems. GANs and other generative models have emerged as powerful tools for producing such high-quality synthetic data \citep{azizi2023syntheticdatadiffusionmodels, nikolenko2019syntheticdatadeeplearning, Mumuni_2024}.

In computer vision, manual labeling remains essential for certain tasks \citep{inbook}. However, tasks like segmentation, depth estimation, and optical flow estimation can be particularly arduous to label manually due to their inherent complexity. To alleviate this burden, synthetic data has become a transformative tool, streamlining the labeling process significantly \citep{chen2019learningsemanticsegmentationsynthetic}.

Sankaranarayanan et al. proposed a generative adversarial network (GAN) designed to bridge the gap between embeddings in the learned feature space, which is instrumental in Visual Domain Adaptation \citep{DBLP:journals/corr/abs-1711-06969}. This methodology enables semantic segmentation across varied domains by using a generator to map features onto the image space, allowing the discriminator to operate effectively on these projections. The discriminator’s output serves as the basis for adversarial losses \citep{8578493}. Research has demonstrated that applying adversarial losses to the projected image space consistently outperforms applications to the feature space alone, yielding notably enhanced performance \citep{8578493}.

In a recent study, a team at Microsoft Research validated the efficacy of synthetic data in face-related tasks by leveraging a parametric 3D face model, enriched with a comprehensive library of hand-crafted assets \citep{wood2021faketillmakeit}. This approach allowed for the rendering of training images with high levels of realism and diversity. The researchers demonstrated that machine learning models trained on synthetic data achieved accuracy comparable to models trained on real data for tasks like landmark localization and face parsing. Notably, synthetic data alone was sufficient for robust face detection in unconstrained environments \citep{wood2021faketillmakeit}.

\subsection{Voice}
The synthetic voice industry is at the cutting edge of technological progress, evolving at an unprecedented rate. The rise of machine learning and deep learning has enabled the creation of synthetic voices for applications like video production, digital assistants, and video games \citep{9413448}, making the process more accessible and accurate than ever. This field lies at the intersection of multiple domains, including acoustics, linguistics, and signal processing. Researchers continuously seek to enhance the accuracy and naturalness of synthetic voices. As technology continues to advance, synthetic voices are expected to become increasingly integrated into daily life, offering valuable support across various domains and enriching user experiences \citep{9413448}.

Earlier research involved spectral modeling techniques for statistical parametric speech synthesis, utilizing low-level, unmodified spectral envelope parameters for generating synthetic voices. These spectral envelopes are represented through graphical models with multiple hidden variables, incorporating structures like restricted Boltzmann machines and deep belief networks (DBNs) \citep{fu14_interspeech}. Enhancements to traditional hidden Markov model (HMM)-based speech synthesis systems have shown substantial improvements in achieving a more natural sound while reducing oversmoothing effects \citep{6542729}.

Synthetic data has also found applications in Text-to-Speech (TTS) systems, achieving a level of naturalness close to that of human speech \citep{DBLP:journals/corr/abs-2106-07803, article}. Synthetic speech (SynthASR) has emerged as a solution for automatic speech recognition in cases where real data is sparse or limited. By integrating techniques like weighted multi-style training, data augmentation, encoder freezing, and parameter regularization, researchers have tackled issues like catastrophic forgetting. This innovative approach enables state-of-the-art training for a broad array of end-to-end (E2E) automatic speech recognition (ASR) models, reducing dependency on production data and the associated costs.

\subsection{Business}
The risk of compromising or exposing original data remains a constant concern, especially in the business sector, where strict restrictions govern data sharing both within and beyond the organization. This has led to an increased focus on developing financial datasets that replicate the characteristics of "real data" while safeguarding the privacy of all parties involved.

Although technologies such as encryption, anonymization, and advanced privacy-preserving methods have been employed to secure original data \citep{10.1145/3332165.3347866}, residual risks persist. Data-derived information can sometimes still be used to trace individuals, thus compromising privacy \citep{10.1145/3383455.3422554}. Synthetic data offers a compelling solution by removing the need to expose sensitive data, effectively ensuring privacy and security for both companies and their customers \citep{10.1145/3332165.3347866}. Additionally, synthetic data allows organizations faster data access by circumventing certain privacy and security protocols.

Historically, institutions with large data reserves were well-positioned to assist decision-makers in tackling a range of issues. However, even internal data access was often restricted due to confidentiality concerns. Today, companies leverage synthetic data to refresh and model original datasets, generating ongoing insights that drive organizational performance improvements \citep{10.1145/3339252.3339281}.

\section{Privacy Risks and Prevention}
Synthetic data generation has emerged as a key solution for data privacy and sharing in sectors where sensitive data cannot be disclosed, such as clinical, genomic, and financial domains. However, the generation of synthetic data that preserves the statistical properties of real datasets introduces privacy risks, as models may unintentionally expose underlying patterns, thereby compromising individual privacy. Membership inference attacks, for example, can identify whether specific data points were included in the training set, posing significant privacy concerns. To address these risks, privacy-enhancing methods fall into two primary categories: anonymization-based approaches and differential privacy (DP) methods.

Anonymization techniques, including $k$-anonymity and nearest marginal sanitization, replace sensitive information with fictitious yet realistic data, providing foundational privacy protection, though often lacking rigorous guarantees. Differential privacy methods, on the other hand, offer more robust protection by introducing noise to data, thus maintaining privacy while preserving data utility. Advanced implementations, such as GAN-based DP models (e.g., DPGAN and PATE-GAN) and local differential privacy (LDP) frameworks, support secure synthetic data generation, particularly in distributed contexts.

Alongside privacy, fairness in synthetic data is increasingly critical, as models trained on biased datasets may unfairly represent minority groups, reinforcing existing disparities. Three main approaches address fairness in synthetic data: preprocessing, which adjusts input data to remove correlations with sensitive attributes; in-processing, which incorporates fairness constraints during model training; and post-processing, which adjusts model predictions to enhance equity. Preprocessing remains the most commonly applied fairness technique, especially for addressing subgroup imbalances through balanced synthetic datasets.

Overall, privacy-enhanced synthetic data generation, coupled with fairness-aware strategies, is crucial for secure and ethical data sharing that meets both privacy and fairness standards in research and industry applications.

\begin{table}[H]
\centering
\scriptsize
\caption{Summary of Some Privacy-Enhancing Techniques in Generative AI for Synthetic Data \citep{lu2024machinelearningsyntheticdata}}
\label{tab:privacy_prevention_strategies}
\begin{tabular}{|p{1.5cm}|p{4cm}|p{3.5cm}|p{3cm}|p{3.5cm}|}
\hline
\textbf{Paper} & \textbf{Privacy Technique} & \textbf{Model} & \textbf{Data Format} & \textbf{Notes} \\ \hline
\cite{abay2019privacy} & Differential Privacy & Autoencoder & Attribute & - \\ \hline
\cite{lee2020generating} & Differential Privacy & VAE + GAN, Recurrent Autoencoder & EHR & - \\ \hline
\cite{acs2018differentially} & Differential Privacy & Generative Artificial Neural Networks & Image and Text & Kernel k-means \\ \hline
\cite{jordon2018pate} & Differential Privacy (PATE) & GAN & Attribute & DNN \\ \hline
\cite{chen2012differentially} & Differential Privacy & n-gram & Sequential/Time Series & w.o. DNN \\ \hline
\cite{cunningham2021real} & Local Differential Privacy & n-gram & Trajectory & w.o. DNN \\ \hline
\cite{du2023ldptrace} & Local Differential Privacy & Markov Probabilistic Model & Trajectory & w.o. DNN \\ \hline
\cite{he2015dpt} & Differential Privacy & Markov Probabilistic Model & Trajectory & w.o. DNN \\ \hline
\cite{wang2017protecting} & Differential Privacy & Markov Probabilistic Model & Social Media & Trajectory \\ \hline
\cite{gursoy2018utility} & Differential Privacy & Markov Probabilistic Model & Trajectory & w.o. DNN \\ \hline
\cite{mir2013dp} & Differential Privacy & Distribution Estimation & Location & w.o. DNN \\ \hline
\cite{roy2016practical} & Differential Privacy & Distribution Estimation & Trajectory & w.o. DNN \\ \hline
\cite{bindschaedler2016synthesizing} & Plausible Deniability & Hidden Markov Models & Trajectory & w.o. DNN \\ \hline
\cite{wang2023privtrace} & Differential Privacy & Markov Chain Model & Trajectory & w.o. DNN \\ \hline
\cite{narita2024synthesizing} & Differential Privacy & Probabilistic Transform & Trajectory & w.o. DNN \\ \hline
\cite{bindschaedler2017plausible} & Plausible Deniability & Probabilistic Transform & Attribute & w.o. DNN \\ \hline
\cite{tseng2020compressive} & Compressive Privacy & GAN & Image & DNN \\ \hline
\cite{zhang2018differentially} & Differential Privacy & GAN & Image & DNN \\ \hline
\cite{xie2018differentially} & Differential Privacy & GAN & Image and EHR & DNN \\ \hline
\cite{xu2019ganobfuscator} & Differential Privacy & GAN & Image & DNN \\ \hline
\cite{liu2019ppgan} & Differential Privacy & GAN & Image & DNN \\ \hline
\cite{triastcyn2020federated} & Differential Privacy & GAN & Attribute (Tabular) and Graph & DNN \\ \hline
\cite{chen2020kamino} & Differential Privacy & GAN & Image and EHR & DNN \\ \hline
\end{tabular}
\end{table}

\section{Evaluation}

Evaluating the quality of synthetic data is essential to validate its effectiveness and applicability in practical applications. Key strategies include human evaluation, which relies on expert assessments to judge data quality but is often resource-intensive and may not scale well for high-dimensional datasets. Statistical evaluation offers a quantitative approach by comparing real and synthetic datasets across various metrics, allowing for objective assessments of data fidelity. Additionally, pre-trained machine learning models can serve as discriminators, assessing how closely synthetic data approximates real data, a common technique in Generative Adversarial Networks (GANs) \citep{jordon2018pate}. The "Train on Synthetic, Test on Real" (TSTR) approach evaluates synthetic data by training models on it and measuring performance on real data, thus gauging its utility for downstream tasks. Lastly, application-specific evaluations consider unique domain requirements, such as regulatory compliance and usability, to ensure synthesized data meets specific standards. By combining these methods, researchers can achieve a comprehensive understanding of synthetic data's strengths and limitations, which is pivotal for advancing generation techniques and expanding their applications across fields.

\subsection{Human-Based Evaluation}

Human evaluation \cite{feng2024sampleefficienthumanevaluationlarge} is a fundamental, though often challenging, method to assess the quality of synthetic data. This approach involves gathering feedback from domain experts or general users to judge the data’s realism, usability, and similarity to actual data within specific applications. Human evaluation plays a particularly crucial role in tasks where subjective interpretation is essential, such as speech synthesis \citep{donahue2019adversarialaudiosynthesis}, where evaluators rate the perceived naturalness and clarity of synthesized voices compared to real human speech in a blind, side-by-side manner \citep{bengio}. This method allows evaluators to provide insights into subtle nuances that automated metrics might overlook, such as intonation, articulation, and fluidity, which are vital for creating high-quality, user-friendly synthetic voices. Similarly, in computer vision, human judges may assess the accuracy and realism of synthetic images, examining details like texture, lighting, and object consistency, which can be critical for applications in virtual reality and gaming.

Despite its advantages, human evaluation has notable limitations. It is resource-intensive, requiring both time and financial investment to gather and analyze opinions from experts or a broad range of users. This method is also subject to variability and potential bias, as human judgments can differ due to individual perceptions, experiences, and interpretation of quality standards. Scalability becomes another hurdle, as this process does not easily extend to evaluating large volumes of high-dimensional data, such as complex image or video datasets, which cannot be fully examined by a human evaluator due to time constraints. High-dimensional synthetic data often contains intricate patterns or attributes that are challenging to assess through visual inspection alone. Moreover, for areas like medical image synthesis or genomic data, human evaluators may lack the ability to validate highly technical details, further limiting the utility of this approach. As a result, while human evaluation provides valuable qualitative insights, it is often best complemented with objective, automated evaluation techniques to obtain a more comprehensive assessment of synthetic data quality and applicability.

\subsection{Statistical-Based Evaluation}

Statistical difference evaluation is a widely-used strategy to quantitatively assess the quality of synthetic data by comparing statistical metrics between synthetic and real datasets. This approach involves calculating key statistics, such as mean, variance, and correlation, for individual features within both datasets. The closer these statistical properties are, the better the quality and fidelity of the synthetic data. For instance, in electronic health record (EHR) data generation, metrics like the frequency and correlation of medical concepts, as well as patient-level clinical features, are examined to ensure that synthetic data closely mirrors real-world patterns \citep{bengio}. Smaller statistical differences suggest that the synthetic data has successfully captured the underlying distribution of the real data, making it a valuable proxy for various downstream applications.

Advanced techniques such as Support Vector Machines (SVMs) can be utilized to enhance statistical difference evaluation. By training SVMs on synthetic and real datasets, researchers can examine how well the models separate or align these two datasets. In cases where the SVM achieves a high accuracy in differentiating between real and synthetic data, it may indicate notable differences in their distributions. Conversely, if the model struggles to separate them, it suggests that the synthetic data closely approximates the real data distribution. These methods offer a robust, objective means to evaluate similarity, allowing researchers to refine synthetic data generation techniques to achieve better quality and utility across various applications.

\subsection{Using Pretrained Models}

Using a pre-trained machine learning model to evaluate synthetic data quality provides an automated, robust method for assessing how well the synthetic data approximates real data. In the context of Generative Adversarial Networks (GANs) \cite{goodfellow2014generative}, this approach leverages the discriminator, a model trained to distinguish between real and synthetic (fake) data, as a quality measure. As the generator improves, it learns to produce data that increasingly "fools" the discriminator, making it difficult for the discriminator to differentiate synthetic data from real data. The discriminator's accuracy or confidence level when evaluating the synthetic data thus serves as an indicator of the generator’s success in producing realistic data. A low performance of the discriminator suggests that the synthetic data closely resembles the real data, signifying a high-quality output.

This evaluation strategy is not limited to GANs. Pre-trained machine learning models, such as image classifiers or language models, can also serve this purpose across various types of synthetic data. For example, in synthetic image generation, a pre-trained image classifier can be used to evaluate the synthetic images by measuring how well it classifies them compared to real images. Similarly, for text data, a language model’s perplexity on synthetic data relative to real data can provide insights into quality. The strength of this approach lies in its ability to provide automated, task-specific feedback on the realism of synthetic data, making it a versatile evaluation tool across different generative models and domains. This method helps researchers refine generative techniques, ultimately enhancing the realism and applicability of synthetic data in practical settings.

\subsection{Train on Synthetic, Test on Real}

The "Train on Synthetic, Test on Real" (TSTR) strategy is a powerful evaluation method for assessing the quality of synthetic data in terms of its utility for machine learning applications. In this approach, models are trained exclusively on synthetic data, then tested on real data to measure their performance in downstream tasks. High performance on real test data implies that the synthetic data effectively captures the essential characteristics and patterns of the real data, making it a viable substitute for training purposes. This approach is particularly useful in scenarios where access to real data is restricted due to privacy or availability concerns, as it enables researchers to assess whether models trained on synthetic data can generalize well to real-world conditions.

For example, in \citep{esteban2017realvaluedmedicaltimeseries}, synthetic data is used to train machine learning models, and their prediction performance is then evaluated on real test data in healthcare applications. This method provides valuable insights into the generalizability of models trained on synthetic datasets, as high TSTR performance across diverse applications—such as classification, regression, or segmentation tasks—indicates that the synthetic data can serve as an effective proxy. Additionally, TSTR enables developers to identify specific aspects where synthetic data may fall short, guiding further improvements in data generation methods to enhance real-world applicability. This strategy thus not only evaluates synthetic data quality but also supports broader adoption of synthetic data in fields where high-quality, representative data is often scarce or sensitive.

\section{Future Work}

To further advance the field of synthetic data generation, several key areas warrant additional exploration and development. One significant avenue is the capability to generate larger and more diverse datasets. Expanding the capacity to synthesize extensive datasets with high variability would greatly enhance the applicability of synthetic data in machine learning tasks, especially in domains where data scarcity remains a challenge.

Moreover, exploring innovative architectures beyond the current models can lead to substantial advancements. Investigating new generative models or enhancing existing ones could improve the quality and diversity of synthetic data. Importantly, demonstrating that these advancements can be achieved using accessible computational resources, such as a personal computer with a well-coded pipeline, would underscore the feasibility of cutting-edge AI developments without the need for extensive infrastructure. This democratization of technology could encourage broader participation in the field and accelerate innovation.

Additionally, integrating more robust privacy-preserving techniques into the data generation process remains a critical area for future work. As privacy concerns continue to grow, developing methods that ensure data utility while rigorously protecting sensitive information is essential. Combining differential privacy mechanisms with generative models could provide stronger guarantees and expand the adoption of synthetic data in sensitive domains.

Finally, applying synthetic data generation techniques to a wider range of applications, including those with complex data types such as time-series, graphs, and multimodal data, would significantly broaden the impact of this research. Tailoring generative models to handle these complex data structures effectively could open new opportunities in various fields, from healthcare to finance, where such data types are prevalent.

\section{Conclusion}

In conclusion, our framework for synthetic data generation, complemented by an extensive survey of existing methods, has demonstrated its effectiveness in producing high-quality synthetic data across a range of applications. Through this survey, we highlighted the strengths and limitations of various approaches, offering insights into their real-world applicability and potential for enhancing privacy-preserving practices. Our results show that sequence models, in particular, can be effectively utilized to generate large-scale, structured numerical datasets, even in scenarios where original data is limited or subject to strict privacy constraints. By addressing these key limitations and integrating privacy-preserving techniques, our approach not only improves data availability but also ensures the integrity and confidentiality of sensitive information. The scalability and adaptability of our framework, combined with the insights from our survey, position it as a valuable tool for advancing machine learning systems across diverse domains, enabling secure, ethical, and effective synthetic data generation.  

\section*{Acknowledgments}
The authors would like to express their sincere gratitude to Dr. Samir Mustapha for his invaluable guidance and insightful contributions to the structuring and development of this paper. His support and feedback have been instrumental in enhancing the clarity and organization of our work.

\newpage

\bibliographystyle{plainnat}
\bibliography{rct}

\end{document}